\documentclass[final]{cvpr}

\usepackage{times}
\usepackage{epsfig}
\usepackage{graphicx}
\usepackage{amsmath}
\usepackage{amssymb}

\usepackage{caption}
\usepackage{csquotes}
\usepackage{amsmath,amsfonts,bm}

\def\eqref#1{equation~\ref{#1}}
\def\1{\bm{1}}

\def\vp{{\bm{p}}}

\def\vw{{\bm{w}}}

\DeclareMathAlphabet{\mathsfit}{\encodingdefault}{\sfdefault}{m}{sl}
\SetMathAlphabet{\mathsfit}{bold}{\encodingdefault}{\sfdefault}{bx}{n}
\newcommand{\tens}[1]{\bm{\mathsfit{#1}}}

\def\tD{{\tens{D}}}

\def\tF{{\tens{F}}}

\def\tI{{\tens{I}}}

\def\tM{{\tens{M}}}

\usepackage{enumitem}
\usepackage[super]{nth}

\newcommand{\crcone}{\raisebox{.5pt}{\textcircled{\raisebox{-.9pt} {1}}}}
\newcommand{\crcthree}{\raisebox{.5pt}{\textcircled{\raisebox{-.9pt} {3}}}}
\newcommand{\crcfour}{\raisebox{.5pt}{\textcircled{\raisebox{-.9pt} {4}}}}
\newcommand{\crcc}{\raisebox{.5pt}{\textcircled{\raisebox{-.5pt} {c}}}}

\usepackage{xcolor,pifont}
\newcommand*\colourcheck[1]{%
  \expandafter\newcommand\csname #1check\endcsname{\textcolor{#1}{\ding{52}}}%
}
\colourcheck{blue}
\colourcheck{green}
\colourcheck{red}

\usepackage{colortbl}

\definecolor{c_muns}{rgb}{0.88,1,1}
\definecolor{c_msup}{rgb}{1.0,0.88,1}
\definecolor{c_data}{rgb}{0.9,0.9,0.9}
\definecolor{c_lowbest}{rgb}{1.0,0.7,0.7}
\definecolor{c_highbest}{rgb}{0.7,0.7,1.0}

\usepackage[pagebackref=true,breaklinks=true,colorlinks,bookmarks=false]{hyperref}

\def\cvprPaperID{8373} % *** Enter the CVPR Paper ID here
\def\confYear{CVPR 2021}

\begin{document}

%%%%%%%%% TITLE
\title{PLADE-Net: Towards Pixel-Level Accuracy for Self-Supervised Single-View Depth Estimation with Neural Positional Encoding and Distilled Matting Loss}

\author{%
  Juan Luis Gonzalez Bello \\
  {\tt\small juanluisgb@kaist.ac.kr}
  % examples of more authors
  \and
  Munchurl Kim \\
  {\tt\small mkimee@kaist.ac.kr}
  \and
  Korea Advanced Institute of Science and Technology \\
}

% \twocolumn[{
% \renewcommand\twocolumn[1][]{#1}
% \maketitle
% \begin{figure*}
%   \centering 
%   \includegraphics[width=1.0\textwidth]{figures/op_img.jpg}
%   \caption{Our proposed FAL-net2 with matting loss and neural positional-encoding estimates depth with pixel-level accuracy.}
%   \label{fig:op_img}
% \end{figure*}
% }]

\twocolumn[{%
\renewcommand\twocolumn[1][]{#1}%
\maketitle
\begin{center}\centering
    \includegraphics[width=1.0\textwidth]{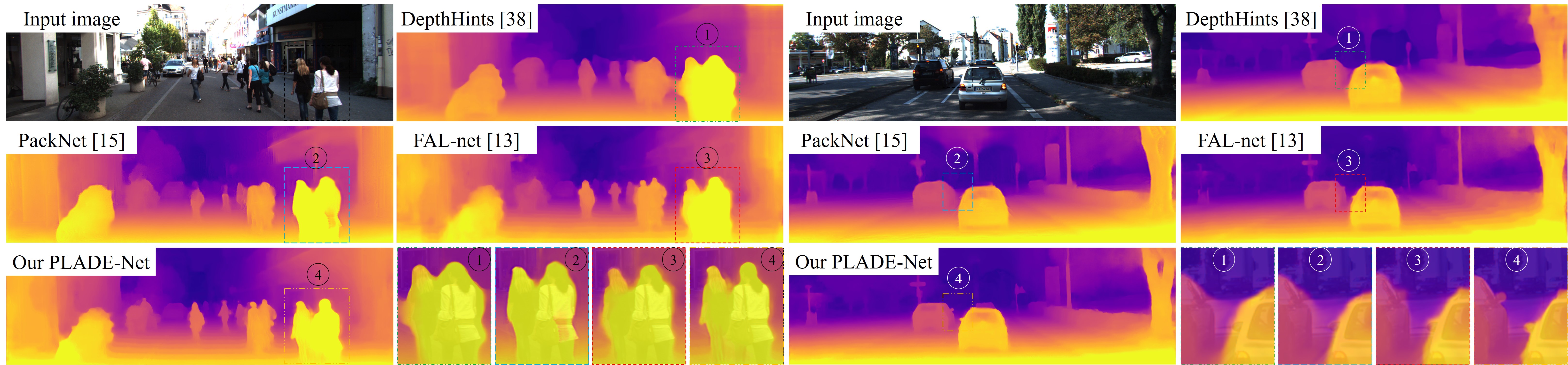}
    \vspace*{-6mm}
	\captionof{figure}{Our proposed PLADE-Net with Neural Positional-Encoding and distilled matting loss estimates depths with pixel-level accuracy.}
	\label{fig:op_img}
\end{center}
}]

% 8 page paper!
%%%%%%%%% ABSTRACT
\begin{abstract}
    \vspace*{-3mm}
    In this paper, we propose a self-supervised single-view pixel-level accurate depth estimation network, called PLADE-Net. The PLADE-Net is the first work that shows unprecedented accuracy levels, exceeding 95\% in terms of the $\delta^1$ metric on the challenging KITTI dataset. Our PLADE-Net is based on a new network architecture with neural positional encoding and a novel loss function that borrows from the closed-form solution of the matting Laplacian to learn pixel-level accurate depth estimation from stereo images. Neural positional encoding allows our PLADE-Net to obtain more consistent depth estimates by letting the network reason about location-specific image properties such as lens and projection distortions. Our novel distilled matting Laplacian loss allows our network to predict sharp depths at object boundaries and more consistent depths in highly homogeneous regions. Our proposed method outperforms all previous self-supervised single-view depth estimation methods by a large margin on the challenging KITTI dataset, with unprecedented levels of accuracy. Furthermore, our PLADE-Net, naively extended for stereo inputs, outperforms the most recent self-supervised stereo methods, even without any advanced blocks like 1D correlations, 3D convolutions, or spatial pyramid pooling. We present extensive ablation studies and experiments that support our method's effectiveness on the KITTI, CityScapes, and Make3D datasets.
\end{abstract}
\vspace*{-3mm}

\section{Introduction}
Recent advances in deep learning have shown state-of-the-art (SOTA) results on the challenging single-view depth estimation (SVDE) and stereo disparity estimation (SDE) tasks. In particular, self-supervised methods for SVDE have reached performance levels similar or even superior to the fully-supervised networks \cite{packing3d, semguide, falnet}. However, the previous SOTA self-supervised methods are unable to predict accurate pixel-level depth estimates, which are often observed along the object's depth boundaries. Predicting pixel-level accurate 3D geometries is essential for robotic grasping, augmented reality, navigation, and 3D object detection. 

In this paper, we present a pixel-level accurate depth estimation network (PLADE-Net) with neural positional encoding and a distilled matting Laplacian loss, both of which allow for consistent depth estimates in homogeneous areas and sharp depth predictions along the object boundaries. Our PLADE-Net outperforms the most recent self-supervised SOTA methods \cite{packing3d,semguide,falnet,unos,reversing} (both mono and stereo) by large margins, achieving an unprecedented accuracy on the challenging KITTI dataset while keeping a low number of parameters. This paper's contributions are:
\begin{enumerate}[leftmargin=*,noitemsep,topsep=0.5pt]

\vspace*{1mm}
\item We propose to exploit and distill the closed-form solution of the matting Laplacian \cite{matting} for self-supervision, leading to a novel loss function that allows for pixel-level accuracy in self-supervised single- and stereo-view DE. 

\vspace*{1mm}
\item We show that neural positional encoding (NPE) can be usefully incorporated into CNNs for depth estimation, as it allows the network to reason about camera distortions, scene orientation, and non-local relationships.

\vspace*{1mm}
\item We present PLADE-Net, a novel network architecture that incorporates NPE. Our PLADE-Net incorporates multi-scale inputs and a single-scale output, opposite to single-scale inputs and multi-scale outputs in previous works \cite{monodepth1,monodepth2,packing3d, depth_hints}. Relative to previous works, our PLADE-Net doubles the number of filter channels in the early feature extraction layers, and halves the number of filter channels in its bottleneck. These seemingly trivial design choices, already make our PLADE-Net, even without our newly proposed loss functions, to outperform the previous SOTA methods.

\vspace*{1mm}
\item The PLADE-Net is the first work that shows unprecedented accuracy levels for SVDE, exceeding 95\% in terms of $\delta^1$ metric on the challenging KITTI\cite{kitti2012} dataset.
\end{enumerate}
\vspace*{1mm}
Figure \ref{fig:op_img} compares the depth estimate performances of the most recent SOTA methods \cite{depth_hints, packing3d, falnet} with respect to our PLADE-Net. As shown in the detailed view of the estimated depth regions (dotted boxes numbered from \crcone\ to \crcfour), our PLADE-Net produces very precisely estimated depths along the object boundaries. Simultaneously, the SOTA methods fail by yielding inaccurate object depths that partially leak into the background. 

Our paper is organized as follows: In Section 2, we review the most relevant self-supervised methods for our work; Section 3 presents our PLADE-Net with neural positional encoding and a distilled matting Laplacian loss with in-depth explanations; In Section 4, we provide extensive ablation studies and experiments that support the effectiveness of our contributions; Finally, we conclude our work in Section 5.

\section{Related Works}
Learning-based self-supervised single view depth estimation (SVDE) is a relatively new problem and has rapidly advanced since it was first proposed in the work of Garg \etal \cite{garg}. Self-supervised SVDE is usually achieved by exploiting the 3D information embedded in datasets that contain multiple captures from the same scene. Previous methods have successfully learned SVDE from stereo pairs \cite{garg, monodepth1, refinedistill, superdepth, net3, depth_hints, infuse_classic, falnet} and video \cite{sfmlearner, monodepth2, packing3d, semguide, depthwild}, by training their CNNs for the backward or forward synthesis of the training image samples, given the target view as input. Similarly, other works \cite{monodepth1, epc, bridging, unos, reversing} have addressed the less ill-posed problem of stereo depth estimation (SDE), where the left and right views are available during training and testing. As our proposed PLADE-Net learns from stereo images, we only review the methods that learn from stereo in this section for the sake of simplicity.

\textbf{Learning SVDE from stereo.} Among the top-performing SVDE methods that learn from stereo we find the works of \cite{infuse_classic, depth_hints, falnet}. The contemporary works of Tosi \etal \cite{infuse_classic} and Watson \etal \cite{depth_hints} proposed to guide the training of their SVDE networks with distilled stereo disparity estimates obtained from the classical approach of semi-global matching (SGM) \cite{sgm0, sgm1}. While Watson \etal \cite{depth_hints} used the SGM disparity as a proxy label when the resulting photometric loss is lower than the CNN-estimated depth, Tosi \etal \cite{infuse_classic} distilled the SGM proxy label via left-right (LR) consistency checks. Inspired by \cite{infuse_classic, depth_hints}, we distill the matting Laplacian with both photometric and LR-consistency checks in this work. 

The recent work of Gonzalez and Kim \cite{falnet}, proposed to \enquote{forget about the LiDAR}, by learning high-quality depths with a multi-view occlusion module and exponentially quantified disparity volumes. Additionally, they proposed a two-stage training strategy to learn view synthesis and refine their network, called FAL-net\cite{falnet}, for SVDE. While their method obtains the SOTA metrics on the KITTI \cite{kitti2012} dataset, their approach is still far from generating pixel-level accurate depths, as shown in Figure \ref{fig:op_img}-\crcthree. Their second stage loss functions cannot enforce sharp object depth boundaries, as they are limited by the computed occlusions' quality, leading to sub-optimal estimates.

\textbf{Learning SDE.} Interestingly, the less ill-posed problem of learning stereo disparity estimation (SDE) in a self-supervised manner has been studied less extensively than the single-view case. The most prominent works include those of Wang \etal \cite{unos} and Aleotti \etal \cite{reversing}. Wang \etal \cite{unos} proposed to exploit spatiotemporal information by learning SDE from stereo videos. Their \enquote{UnOS} learns optical flow, stereo disparity, and camera pose by spatially and temporally projecting the target views into the spatiotemporal reference images and measuring the reconstruction errors to provide means of self-supervision. The work of Aleotti \etal proposes to distill the disparity estimates from a monocular disparity competition network to provide additional proxy labels for the SDE task. Aleotti \etal achieve the SOTA by training existing networks \cite{monodepth2, piramid_sm} with their monocular proxy labels, which remove the well-known stereo artifacts caused by occlusions \cite{monodepth1, sgm0, sgm1, reversing}. 
 
\section{Method}
We propose a novel Pixel-Level Accurate Depth Estimation network, called PLADE-Net, with neural positional encoding and a distilled matting Laplacian loss. Architecture-wise, neural positional encoding is incorporated into our PLADE-Net to learn location-specific image features.  Training-wise, our PLADE-Net learns single-view depth from stereo pairs in a two-stage training strategy following the previous work \cite{falnet}. In the first stage of training, our PLADE-Net is trained for simple stereoscopic view synthesis with a combination of $l_1$, perceptual \cite{perceptual}, and smoothness losses. In the second stage, our network is fine-tuned with an occlusions-free reconstruction loss with the multi-view occlusion module and other secondary smoothness and mirror losses defined in \cite{falnet}. More importantly, a distilled matting Laplacian loss is newly proposed, allowing the learning of highly accurate pixel-level depth estimates.

In the following subsections, we introduce an image (and inverse depth) formation model (which follows the one defined in \cite{falnet}) and describe the intuition behind our main contributions for \textit{neural positional encoding} and \textit{a distilled matting Laplacian loss}. We then describe the details of our network architecture and training loss functions in Subsections 3.4 and 3.5, respectively.

\subsection{Stereoscopic Image Formation Model}
We build our PLADE-Net based on the work of Gonzalez and Kim \cite{falnet}, as their method showed SOTA results for learning single-view depth from stereo images. Therefore, we adopt their image formation model in which the convolutional neural network (CNN) outputs a disparity probability logit volume $\tD^L_L$ for a given left input view $\tI_L$. $\tD^L_L$ can be either progressively projected to the right-view and soft-maxed channel-wise to form the right-from-left disparity probability volume $\tD^{PR}_L$ or simply soft-maxed to generate the left disparity probability volume $\tD^{PL}_L$. $\tD^{PR}_L$ can be used for stereoscopic view synthesis by
\begin{equation} \label{eq:synth_right}
\tI'_R = \textstyle \sum_{n=0}^{N} g\left(\tI_L, d_n\right) \odot \tD^{PR}_{L_n},
\end{equation}
where $\odot$ indicates the Hadamard product, $g(\cdot)$ denotes a shifting of the input image to the left by $d_n$ pixels, and $N$ is the number of planes in the probability volume $\tD^{PR}_L$. On the other hand, $\tD^{PL}_L$ can be used to extract the disparity map $\tD'_L$, which is learned as a by-product from the view synthesis task, as defined by
\begin{equation} \label{eq:disp}
\tD'_L = \textstyle \sum_{n=0}^{N} d_n \tD^{PL}_{L_n}
\end{equation}
We also adopt the exponential disparity quantization in \cite{falnet}, as it is well-posed for the depth estimation task. Exponential quantization takes into account the inverse relationship between disparity and depth by distributing far- and close-by quantization levels more evenly and is given by
\begin{equation} \label{eq:exp_disc}
    d_n = d_{max}e^{\ln{d_{max}/d_{min}}(n/N-1)},
\end{equation}
where $d_{max}$ and $d_{min}$ are the minimum and maximum disparity hyper-parameters. For a fair comparison with \cite{falnet}, we set $d_{max}=300$ and $d_{min}=2$ for all our experiments.

\subsection{Neural Positional Encoding}
\begin{figure}
  \centering 
  \includegraphics[width=0.47\textwidth]{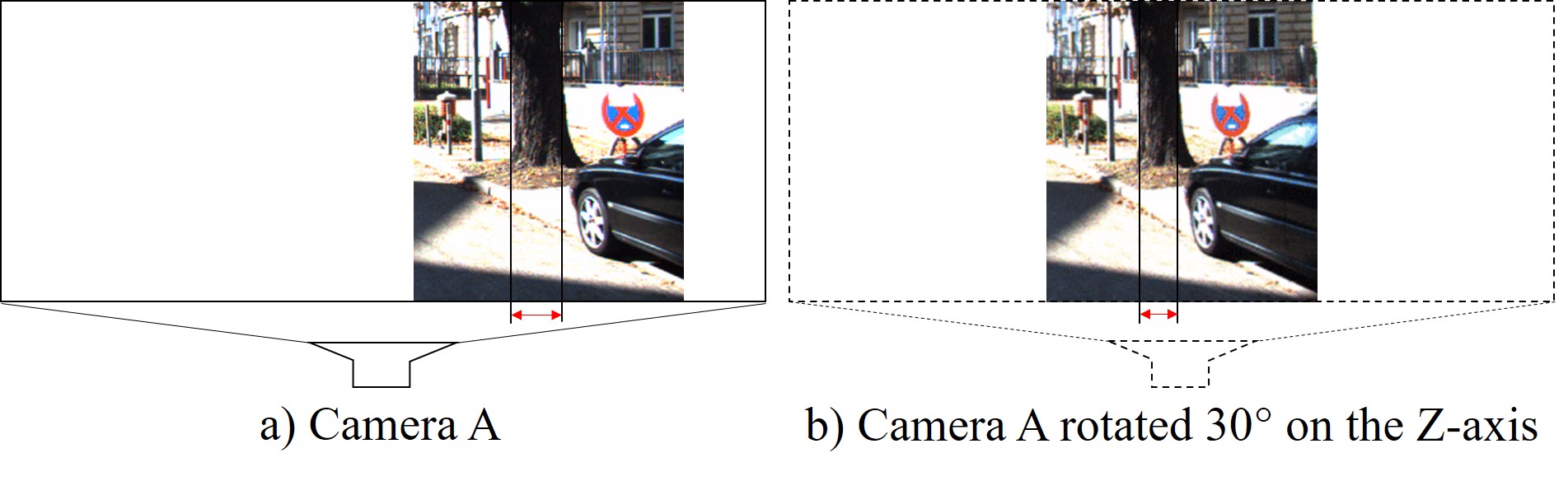}
  \vspace*{-3mm}
  \caption{An illustration of projection distortions in image borders. Objects closer to the image borders appear stretched and closer, like the three in (a) in comparison with the three in (b).}
  \label{fig:rot_effect}
\end{figure}

It is well-known that convolutional neural networks on their own are very well capable of encoding positional information \cite{how_much_pos}. However, since the local CNN filters are shared across spatial locations, a network, trained with randomly cropped patches from the original image data, will struggle to learn location-specific features, such as lens or projection distortions, ground versus sky regions, and potentially non-local relationships. In particular, lens and projection distortions make the objects near image borders appear to be more stretched than those in the image centers, the degree of which often depends on the camera focal length. This is a potential source of confusion for the CNNs, as two objects in the same distance to a camera will be projected differently on the camera plane, depending on their relative position to the resulting image, as illustrated in Figure \ref{fig:rot_effect}.

The relative object size is an essential cue for depth estimation. It can affect the estimation accuracy if a network does not have a means of understanding the locations of the training patches in their original images. To provide the network with a mechanism to account for the likelihood of objects being stretched when they are located close to the image borders, we propose neural positional encoding (NPE) for depth estimation. We realize NPE into our PLADE-Net as the concatenation of deep positional features at each encoder stage. A deep positional feature map $\tF_{npe}$ is obtained by processing the pixel location $\vp=(x,y)$ information of each patch with two fully-connected layers with exponential linear unit (ELU) activations, which is given by:
\begin{equation} \label{eq:npe}
\tF_{npe}(\vp) = elu(\vw_2 \cdot elu(\vw_1 \cdot \vp + b_1) + b_2),
\end{equation}
where $w_{1,2}$ and $b_{1,2}$ are the learnable weights and biases of our neural positional encoding layers. The operation in Eq. \ref{eq:npe} can be trivially realized with $1\times1$ convolutions in available deep learning libraries. Note that in contrast with \cite{camconvs}, we do not concatenate $x$- and $y$-coordinates of $\vp$, but do concatenate our deep positional feature maps into the downstream convolutional layers in our PLADE-Net. It should be noted again that, in our neural positional encoding, $\vp=(x,y)$ are the pixel locations of the patches relative to their original images before cropping.

\begin{figure}
  \centering 
  \includegraphics[width=0.48\textwidth]{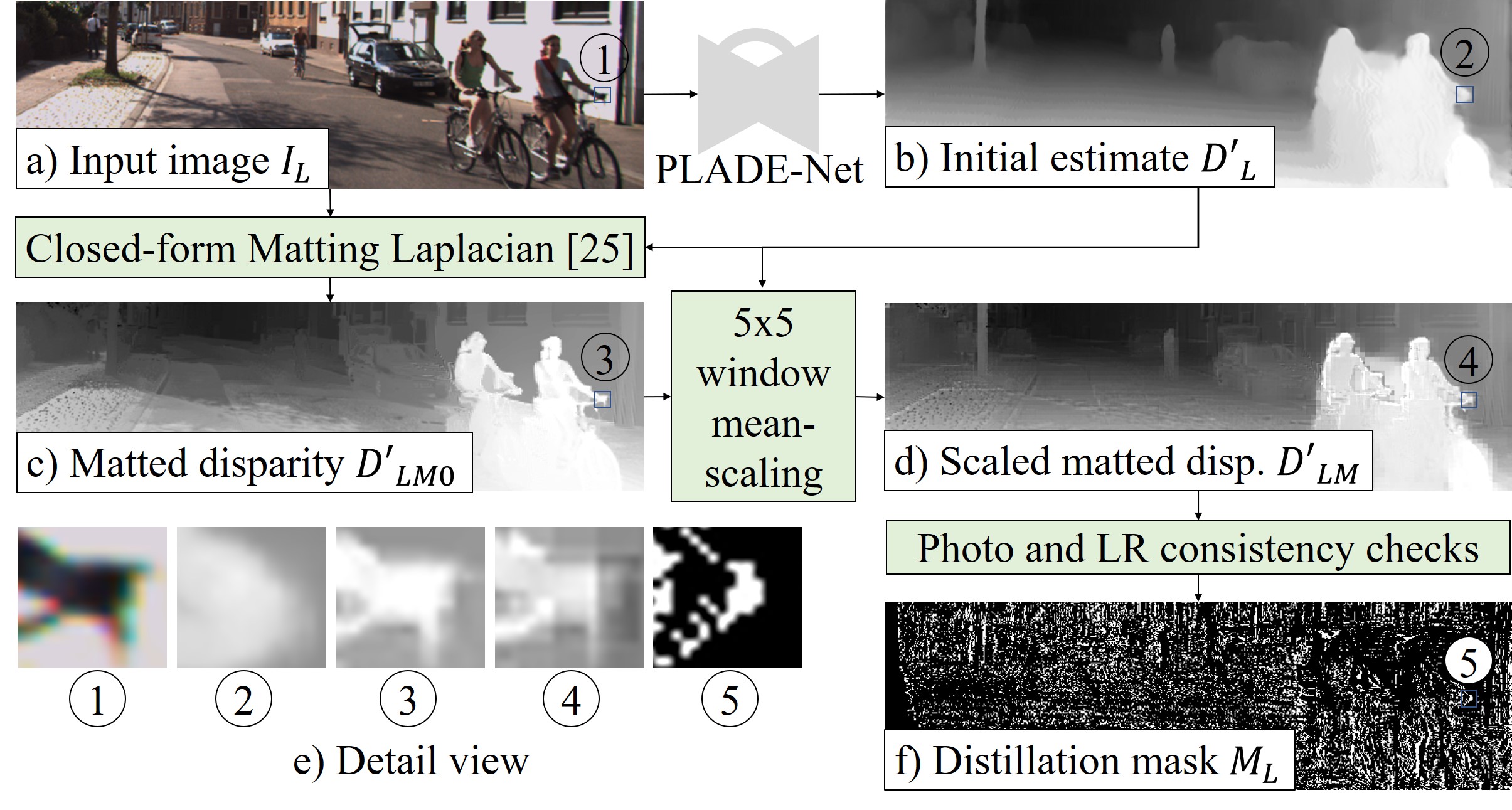}
  \vspace*{-7mm}
  \caption{Matted disparity distillation process.}
  \label{fig:matting_dist}
  \vspace*{-4mm}
\end{figure}

\subsection{Distilling the Matting Laplacian}
The closed-form solution to the matting Laplacian \cite{matting} is a useful tool in classical low-level computer vision. It can sharply segment an input image based on pixel intensity and proximity in the input image, a roughly estimated or user-defined segmentation map, and a confidence map. The matting Laplacian has been used to refine depth estimates, generating structurally sharp but incorrectly labeled matted depth maps. We exploit the strong features of the matting Laplacian to learn highly accurate pixel-level depth in a self-supervised fashion, remedying its weak points by distilling the matted depth maps with photometric and left-right consistency \cite{monodepth1} checks.

Our matting Laplacian distillation process is depicted in Figure \ref{fig:matting_dist}. Given an input image $\tI_L$ and its initial depth estimate $\tD'_{L0}$ by our PLADE-Net, we generate a matted disparity $\tD'_{LM0}$ following \cite{matting}. As can be observed in Figure \ref{fig:matting_dist}-(c), $\tD'_{LM0}$ is very sharp, but many pixels are wrongly labeled, such as the tree at the left-hand side of $\tI_L$, the biker in the background, and the overall road depths. To remedy this, we apply a $5 \times 5$ local window mean scaling to obtain the locally scaled matted disparity map $\tD'_{LM}$, as shown in Figure \ref{fig:matting_dist}-(d). Then, we estimate a distillation mask $\tM_L$ via photometric and left-right consistency checks as given by:
\begin{equation} \label{eq:dist_md}
\begin{split}
\tM&_L = \big[\big|\tI_L - g(\tI_R,\tD'_{LM})\big| < \big|\tI_L - g(\tI_R,\tD'_L)\big|\big] \odot \\
&\big[\big|\tD'_{LM} - g(\tD'_{RM},\tD'_{LM})\big| < \big|\tD'_L - g(\tD'_R,\tD'_L)\big|\big],
\end{split}
\end{equation}
where $\tD'_{R0}$ and $\tD'_{RM}$ are the initial disparity estimate and the locally scaled matted disparity map for the corresponding right-view input image $\tI_R$, respectively. $g(\cdot)$ works as a backward-warping operation. In Eq. \ref{eq:dist_md}, if the inequalities in both brackets are satisfied at a pixel location, the resulting mask value at that location is 1, otherwise 0. The distillation map for the right view $\tM_R$ is obtained by swapping the L and R sub-scripts in Eq. \ref{eq:dist_md}. Eq. \ref{eq:dist_md} selects as a source for self-supervisions the pixel depths in $\tD'_{LM}$ that (i) generate better backward warped images and (ii) are more consistent with their corresponding right view pixel depths. As can be noted in the detailed view of the hand of the biker in Figure \ref{fig:matting_dist}-(e), the matted disparity becomes dramatically sharper than the initial disparity estimate but has incorrect values. The mean-scaled matted disparity is both sharper and correctly scaled. As expected, the distillation mask is active on the biker's hand edges, which guides our network to generate pixel-level accurate depth estimates.

\subsection{Network Architecture}
\begin{figure}
  \centering 
  \includegraphics[width=0.48\textwidth]{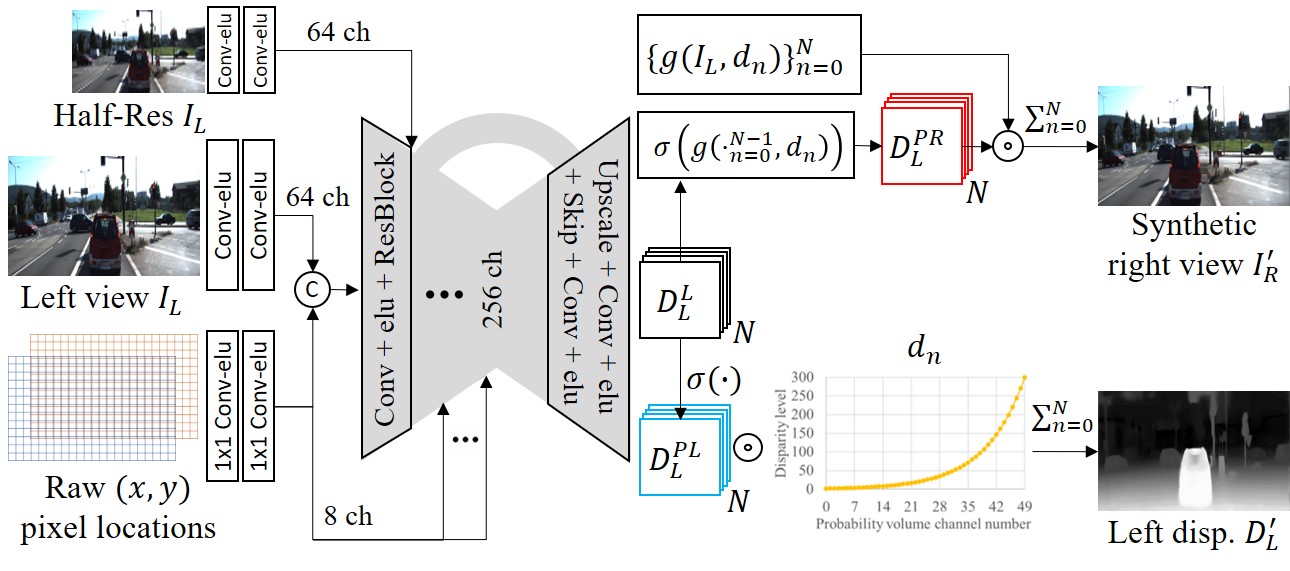}
  \vspace*{-5mm}
  \caption{High-level overview of our proposed PLADE-Net.}
  \label{fig:network}
  \vspace*{-4mm}
\end{figure}

Our proposed PLADE-Net adopts the simple auto-encoder backbone from \cite{falnet}, with residual blocks in the encoder side and nearest-upscale-based up-convolutions followed by skip-connections in the decoder side. However, we considerably change the learned feature maps' distribution by doubling the extracted features in the shallow convolutional layers (from 32 to 64) and halving the number of feature maps in the bottleneck (from 512 to 256). Our PLADE-Net is depicted in Figure \ref{fig:network} and incorporates our proposed NPE by concatenating (denoted by \crcc) deep positional features at each encoder stage's input. 

Contrary to the previous works \cite{monodepth1, monodepth2, depth_hints, packing3d, falnet} that incorporate a single-scale input and multi-scale outputs, our PLADE-Net adopts multi-scale inputs and a single scale output. In our PLADE-Net, low-level features are extracted from a bilinearly downscaled version of the input image $\tI_L$ and concatenated into the second encoder stage's input, as depicted to the left-hand side of Figure \ref{fig:network}. Our PLADE-Net outputs a single-scale disparity logit volume $\tD^L_L$, which can be employed for novel view synthesis and single-view depth estimation, as shown to the right-hand side of Figure \ref{fig:network}, according to Eqs. \ref{eq:synth_right} and \ref{eq:disp}, respectively.

Our PLADE-Net delivers higher performance with an equal or lower number of parameters than the previous works, with 15M versus 17M of the previous SOTA \cite{falnet}. It is worth noting that our PLADE-Net achieves the SOTA performance without the need for any advanced layer such as attention, batch/group normalization, sub-pixel convolution, or spatial pyramid pooling. Detailed architecture layer information can be found in the supplementary materials.

\subsection{Loss Functions}
Following the training strategy in \cite{falnet}, we train our PLADE-Net in two stages. In the first stage, we focus on learning stereoscopic view synthesis, which can be understood as training the top output branch of our PLADE-Net in Figure \ref{fig:network}, which generates a synthetic right view. In the second stage, we train our PLADE-Net with an occlusions-free reconstruction loss. Still, more importantly, we incorporate additional loss functions that affect the lower output branch in Figure \ref{fig:network}, which generates a near pixel-accurate disparity estimate. 
The total loss function ($l_{s1}$) in the first stage of training is a combination of $l_1$, perceptual \cite{perceptual} ($l_p$), and disparity smoothness ($l_{ds}$) losses, as given by 
\begin{equation} \label{eq:l_s1}
l_{s1} = l_1 + \alpha_pl_p + \alpha_{ds}l_{ds},
\end{equation}
where $\alpha_p$ and $\alpha_{ds}$ are empirically set to 0.01 and 0.0004, respectively, to balance their contributions. The total loss $l_{s2}$ for the second training stage adds our novel distilled matting Laplacian loss ($l_{dm}$), a deep corr-$l_1$ loss ($l_{dc}$), and the mirror loss ($l_{m}$) in \cite{falnet} to $l_{s1}$, and is defined by:
\begin{equation} \label{eq:l_s2}
l_{s2} = l_{s1} + l_{m} + \alpha_{dm}l_{dm} + \alpha_{dc}l_{dc},
\end{equation}
where $\alpha_{dm}=0.25$ and $\alpha_{dc}=0.01$ are empirically set to weight the contributions of the distilled matting Laplacian and the deep corr-$l_1$ losses, respectively. 

In the first training stage, $l_{s1}$ is computed for the left view only. In the second stage, $l_{s2}$ is computed for both left and right views, giving the actual total loss of $l=(l^L_{s2}+l^R_{s2})/2$. For the sake of completeness, we describe $l_{s1}$ and $l_{m}$ in detail in our supplemental materials.

\vspace*{1mm}
\textbf{Deep Corr-$l_1$ Loss.} Inspired by \cite{structurepre}, we explored training our PLADE-Net with a deep corr-$l_1$ loss, to encourage the generation of depth estimates with structural details similar to the ones in the single-view input. However, we observed marginal performance improvements and depth artifacts, as further shown in Section 4. Nevertheless, we observed an affinity between the deep corr-$l_1$ loss and our proposed distilled matting Laplacian loss. Our deep corr-$l_1$ loss ($l_{dc}$) penalizes the deep-auto-correlation difference between the input image and the predicted depth map. Deep-auto-correlation is obtained by measuring the auto-correlation of the deep features of $\tI_L$ or $\tD'_L$, extracted by the third maxpool layer of a pre-trained VGG19 \cite{vgg} for the image classification task, denoted by $\phi(\cdot)$. $l_{dc}$ is then given by:
\begin{equation} \label{eq:dcl_loss}
l_{dc} = ||\text{acorr}(\phi(\tD'_L), k)-\text{acorr}(\phi(\tI_L), k)||_1,
\end{equation}
where $\text{acorr}(\cdot, k)$ is the auto-correlation operator on a $k$$\times$$k$ window, empirically set to $k=3$ in all our experiments.

\vspace*{1mm}
\textbf{Distilled Matting Laplacian Loss.} We previously detailed our matting Laplacian distillation process in Subsection 3.3. Given the locally scaled matted left disparity map $\tD'_{LM}$ and distillation mask $\tM_L$, our distilled matting Laplacian loss $l_{dm}$ is simply given by:
\begin{equation} \label{eq:dm_loss}
l_{dm} = (1 / \text{max}(\tD'_L))||\tM_L \odot(\tD'_L - \tD'_{LM})||_1,
\end{equation}
where $\text{max}(\tD'_L)$ is the maximum disparity value in the scene and normalizes the loss between 0 and 1. By incorporating $\tM_L$ into Eq. (\ref{eq:dm_loss}), we can keep the highly detailed matted depths while filtering out the incorrectly labeled pixels commonly present in image matting. 

\begin{table}[t]
    \small
    \centering
    \setlength{\tabcolsep}{2.5pt}
    \begin{tabular}{lcccccc}
\hline
Methods & Data & abs rel\cellcolor{c_lowbest} & sq rel\cellcolor{c_lowbest} & rmse\cellcolor{c_lowbest} & rmse$_{log}$\cellcolor{c_lowbest}  & $\delta^1$\cellcolor{c_highbest}  \\ 
\hline

\multicolumn{7}{l}{Effects of NPE in the \nth{1} stage of training}\\
NPE & K+CS & 0.070 & 0.291 & 2.910 & 0.107 & 0.942\\
NPE & K & 0.071 & 0.318 & 3.236 & 0.113 & 0.934 \\
PE  & K+CS & 0.070 & 0.292 & 2.988 & 0.109 & 0.939 \\
% PE-1 & K+CS & 0.070 & 0.284 & 2.888 & 0.107 & 0.940 \\
w/o PE & K+CS & 0.075 & 0.317 & 2.990 & 0.111 & 0.938\\ % & 0.989 & 0.997\\
\hline

\multicolumn{7}{l}{Effects of NPE in the \nth{2} stage of training ($a_{dc}$=0.01, $a_{dm}$=0.25)}\\
NPE & K+CS & 0.066 & 0.263 & 2.726 & 0.102 & 0.949 \\
NPE & K & 0.066 & 0.274 & 2.881 & 0.105 & 0.944 \\ % & 0.992 & 0.998\\
PE  & K+CS & 0.067 & 0.268 & 2.797 & 0.104 & 0.945 \\		
% PE-1 & K+CS & 0.066 & 0.264 & 2.719 & 0.102 & 0.947 \\		
w/o PE & K+CS & 0.074 & 0.298 & 2.842 & 0.108 & 0.942\\ % & 0.989 & 0.998 \\		
\hline

\multicolumn{7}{l}{Effects of $l_{dc}$ and $l_{dm}$ in the \nth{2} stage of training}\\
$a_{dc}$=0, $a_{dm}$=0.25 & K+CS & 0.068 & 0.268 & 2.741 & 0.103 & 0.948 \\
% 0.0675    0.2667    2.7406    0.1032    0.9475    0.9914    0.9979 nPE
$a_{dc}$=0.01, $a_{dm}$=0 & K+CS & 0.067 & 0.267 & 2.775 & 0.104 & 0.945 \\
% 0.0669    0.2662    2.7650    0.1033    0.9471    0.9911    0.9979 nPE (right)
% 0.0673    0.2781    2.8696    0.1093    0.9436    0.9897    0.9972 nPE
% 0.0666    0.2665    2.7745    0.1043    0.9451    0.9916    0.9980 PE-1
$a_{dc}$=0, $a_{dm}$=0 & K+CS & 0.067 & 0.270 & 2.777 & 0.104 & 0.943 \\
% 0.0661    0.2627    2.7884    0.1033    0.9466    0.9913    0.9979 nPE
% 0.0667    0.2608    2.7474    0.1030    0.9471    0.9916    0.9981
% 0.0670    0.2701    2.7774    0.1043    0.9433    0.9918    0.9980 PE-1
\hline
FAL-net \cite{falnet} & K+CS & 0.071 & 0.287 & 2.905 & 0.109 & 0.941 \\
\hline

\hline
\end{tabular}
    \vspace*{-2mm}
    \caption{Ablation studies of our PLADE-Net on KITTI\cite{kitti2012}. Metrics are \colorbox{c_lowbest}{the lower the better} and \colorbox{c_highbest}{the higher the better}.}
    \label{tab:ablation}
    \vspace*{-2mm}
\end{table}

\begin{figure}
  \centering 
  \includegraphics[width=0.48\textwidth]{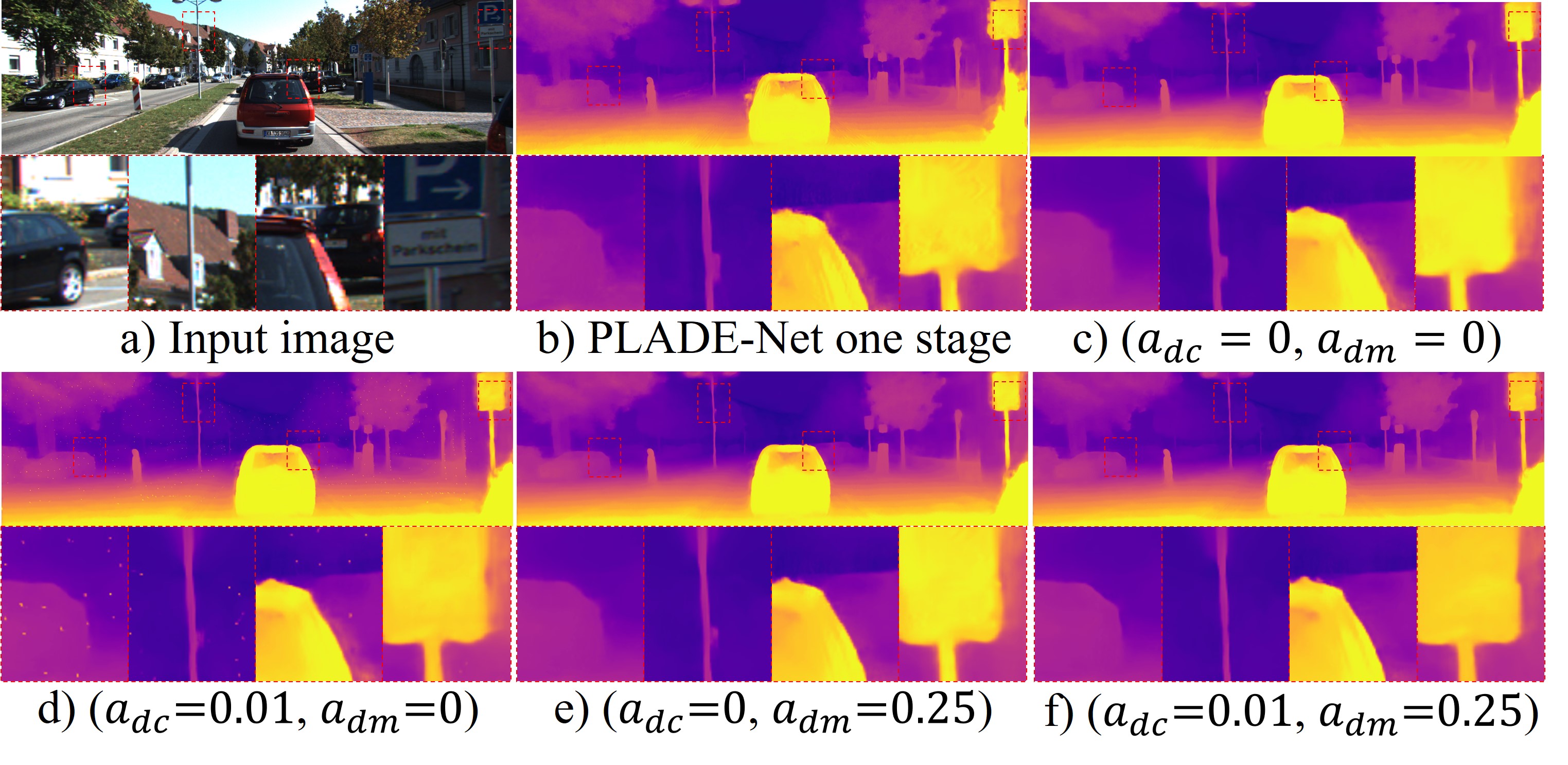}
  \vspace*{-8mm}
  \caption{Ablation studies on our distilled matting Laplacian loss.}
  \label{fig:loss_ablation}
  \vspace*{-3mm}
\end{figure}

\section{Experiments and Results}
% We evaluate our approach with extensive experiments on the KITTI \cite{kitti2012}, CityScapes \cite{cityscapes}, and Make3D \cite{make_3d} datasets, and compare our PLADE-Net with the most recent SOTA methods. Additionally, we provide intensive ablation studies supporting the effectiveness of our proposed NPE and distilled matting Laplacian loss.

\subsection{Datasets}
\textbf{KITTI\cite{kitti2012}.} To compare with a wider spectrum of recent works, we utilize the Eigen train split \cite{eigen} (K), which is a subset of the KITTI \cite{kitti2012} training set, consisting of 22,600 left-right training image pairs captured from a moving car. Following the standard practice, we test our method on the KITTI Eigen test split in its original \cite{eigen} and improved \cite{kitti_official} versions, which contain 697 and 652 images with projected LiDAR ground truths (GT), respectively. The improved Eigen test split contains denser GTs by selectively accumulating LiDAR points from 5 consecutive frames. Performance is measured with the metrics defined in \cite{eigen} (up to 80m). Additionally, in our experiments, we propose a \enquote{naive} stereo input extension of our PLADE-Net, which is trained with a split obtained from the intersection of the KITTI Eigen train set \cite{eigen} and the KITTI Split \cite{monodepth1}. The resulting Stereo-Split excludes scenes from the KITTI Eigen test split \cite{eigen} and the KITTI2015 training set \cite{kitti2015}. The KITTI2015 \cite{kitti2015} training set consists of 200 image pairs with CAD-refined LiDAR GT and is the default benchmark to evaluate self-supervised stereo networks.

\textbf{CityScapes\cite{cityscapes}.} In most of our ablation studies, we concurrently train the PLADE-Net variations with the CityScapes \cite{cityscapes} dataset (following the multi-dataset training procedure in \cite{deep3dpan, falnet}) to ensure that they do not under-perform due to the lack of enough data. The CityScapes \cite{cityscapes} dataset consists of 24,500 stereo pairs without depth GTs, and similar to KITTI \cite{kitti2012}, it is captured from a driving perspective. We follow the car hood and border artifacts removal procedures from \cite{monodepth1, deep3dpan, falnet}.

\textbf{Make3D\cite{make_3d}.} To test the generalization power of our PLADE-Net, we evaluate it on the Make3D \cite{make_3d} Test134 dataset, which is made of 134 high-resolution RGB outdoor images with low-resolution depth GTs. We followed the evaluation procedure defined in \cite{monodepth1}, with the C1 metrics (up to 70m) defined in \cite{liudisc}.

\subsection{Implementation Details}
Following the training procedure in \cite{falnet}, we trained our PLADE-Net for 50 epochs in the first training stage and 10 epochs in the second stage by the Adam\cite{adam} optimizer with an initial learning rate of 0.0001 and 0.00005 for the first and second training stages, respectively. The learning rate was reduced to half at epochs [30, 40, 50] in the first training stage and epochs [5, 10] in the second stage. Data augmentations on-the-fly were incorporated into our network training. For a fair comparison with previous works, we adopted random resizing from 0.75 to 1.5, followed by $192\times640$ random cropping, random left-right flipping, random gamma, random brightness, and random individual color brightness. Inference was run at full image resolution.

\subsubsection{Computing the Matting Laplacian}
Computing the closed-form solution to the matting Laplacian \cite{matting} is expensive, taking up to 30 (60) seconds in matting a KITTI \cite{kitti2012} (CityScapes \cite{cityscapes}) depth-RGB sample. For this reason, we first generated a matted disparity complementary dataset instead of matting on-the-fly. We used our PLADE-Net with one stage of training to build such a matted disparity dataset. We procured to apply the corresponding spatial data augmentations (resizes, crops, and flips) to the matted disparity samples during training. Our distillation process re-scales the matted disparity values, thus not requiring to apply scaling factors during data sampling. 
% We will make our matted disparity complementary datasets for KITTI\cite{kitti2012} and CityScapes\cite{cityscapes} publicly available.

\begin{figure}
  \centering 
  \includegraphics[width=0.48\textwidth]{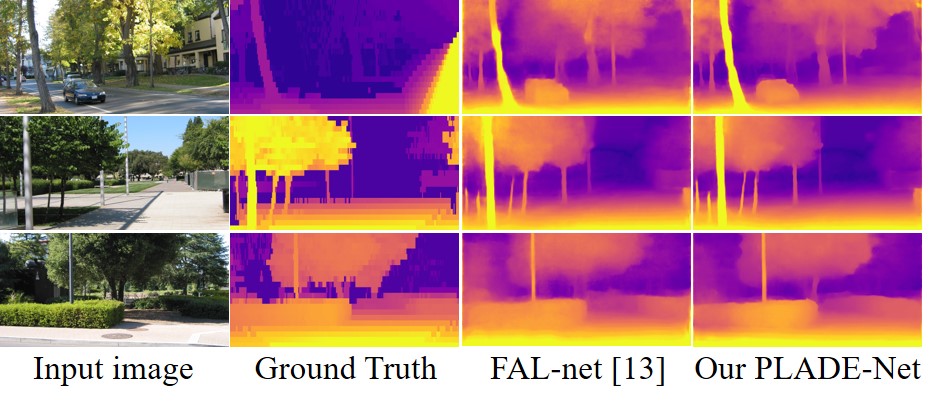}
  \vspace*{-9mm}
  \caption{Qualitative comparisons on the Make3D dataset \cite{make_3d}.}
  \label{fig:make_3d}
  \vspace*{-2mm}
\end{figure}

\begin{table}
    \small
    \centering
    \setlength{\tabcolsep}{5.5pt}
    % \begin{tabular}{lccccc|lccccc}
% \hline
% Method & Sup & abs rel & sq rel  & rms  & Avg $log_{10}$ &
% Method & Sup & abs rel  & sq rel  & rms  & Avg $log_{10}$\\  
% \hline
% Liu \etal \cite{liudisc}       & D & 0.475 & 6.562 & 10.05 & 0.165 & 
% Laina \etal \cite{lainadeeper} & D & \textbf{0.204} & \textbf{1.840} & \textbf{5.683} & \textbf{0.084} \\
% \hline
% SFMLearner \cite{sfmlearner}   & V & 0.383 & 5.321 & 10.47 & 0.478 &
% Monodepth (PP) \cite{monodepth1} & S & 0.443 & 7.112 & 8.860 & 0.142 \\
% Monodepth2 \cite{monodepth2} & V &0.322 & 3.589 & 7.417 & 0.163 &
% Wang \etal \cite{wangdirect}   & S & 0.387 & 4.720 & 8.090  & 0.204 \\	
% Zhou \etal \cite{highresdepth}  & V & 0.318 & 2.288 & 6.669 & - &
% Glez. and Kim \cite{gonzalezdisp} & S & 0.323 & 4.021 & 7.507 & 0.121 \\ 
% FAL-netB33                   & S & 0.317 & 3.265 & 7.053 & 0.121 &	
% FAL-netB33 (PP)              & S & 0.306 & 2.979 & 6.771 & 0.118 \\	
% FAL-netB49                   & S & \textul{0.256} & \textul{2.179} & \textul{6.201} & \textul{0.106} &
% FAL-netB49 (PP)              & S & \textbf{0.254} & \textbf{2.140} & \textbf{6.139} & \textbf{0.105} \\
% \hline
% \end{tabular}

\begin{tabular}{lccccc}
\hline
Method & Sup & Data & abs rel\cellcolor{c_lowbest} & sq rel\cellcolor{c_lowbest}  & rmse\cellcolor{c_lowbest} \\
\hline

Liu \etal \cite{liudisc}       & D & M3D & 0.475 & 6.562 & 10.05 \\
Laina \etal \cite{lainadeeper} & D & M3D & \textbf{0.204} & \textbf{1.840} & \textbf{5.683} \\
\hline

% SFMLearner \cite{sfmlearner}   & V & K & 0.383 & 5.321 & 10.47 \\
% Monodepth (PP) \cite{monodepth1}    & S & CS & 0.443 & 7.112 & 8.860 \\
Monodepth2 \cite{monodepth2}   & V & K & 0.322 & 3.589 & 7.417 \\
Wang \etal \cite{wangdirect}   & S & K & 0.387 & 4.720 & 8.090 \\	
Glez. and Kim \cite{gonzalezdisp} & S & K & 0.323 & 4.021 & 7.507 \\ 
Zhou \etal \cite{highresdepth} & V & K & 0.318 & 2.288 & 6.669 \\
% FAL-net \cite{falnet}          & S & K & 0.297 & 2.913 & 6.810 \\
FAL-net \cite{falnet} (PP)     & S & K & 0.284 & 2.803 & 6.643 \\
% FAL-net \cite{falnet}          & S & K+CS & 0.256 & 2.179 & 6.201 \\
FAL-net \cite{falnet} (PP)     & S & K+CS & \underline{0.254} & \underline{2.140} & 6.139 \\
% PLADE-Net                       & S & K+CS & 0.262 & 2.263 & 6.196 \\
% 0.2617    2.2634    6.1955    0.1073    0.6187    0.8429    0.9279
% PLADE-Net (PP)                  & S & K+CS & 0.257 & 2.215 & \textbf{6.048} \\
% 0.2568    2.2150    6.0480    0.1045    0.6227    0.8557    0.9336

PLADE-Net                       & S & K & 0.276 & 2.635 & 6.546 \\
    % 0.2757    2.6348    6.5458    0.1068    0.6313    0.8432    0.9269
PLADE-Net (PP)                  & S & K & 0.265 & 2.469 & 6.373 \\
    % 0.2652    2.4691    6.3729    0.1039    0.6412    0.8510    0.9298

PLADE-Net                       & S & K+CS & 0.257 & 2.146 & \underline{6.097} \\
% 0.2571    2.1464    6.0968    0.1051    0.6275    0.8495    0.9302
PLADE-Net (PP)                  & S & K+CS & \textbf{0.253} & \textbf{2.100} & \textbf{6.031} \\
% 0.2525    2.1003    6.0312    0.1033    0.6327    0.8559    0.9351

\hline
\end{tabular}

% These are C1 errors, that is on areas less than 70m			
% These metrics are defined in Discrete-Continuous Depth Estimation from a Single Image
			
    \vspace*{-2mm}
    \caption{Results on Make3D \cite{make_3d}. All self-supervised methods benefit from median scaling. M3D: Training on the Make3D\cite{make_3d}.}
    \label{tab:make3d}
    \vspace*{-2mm}
\end{table}

\begin{figure}
  \centering 
  \includegraphics[width=0.48\textwidth]{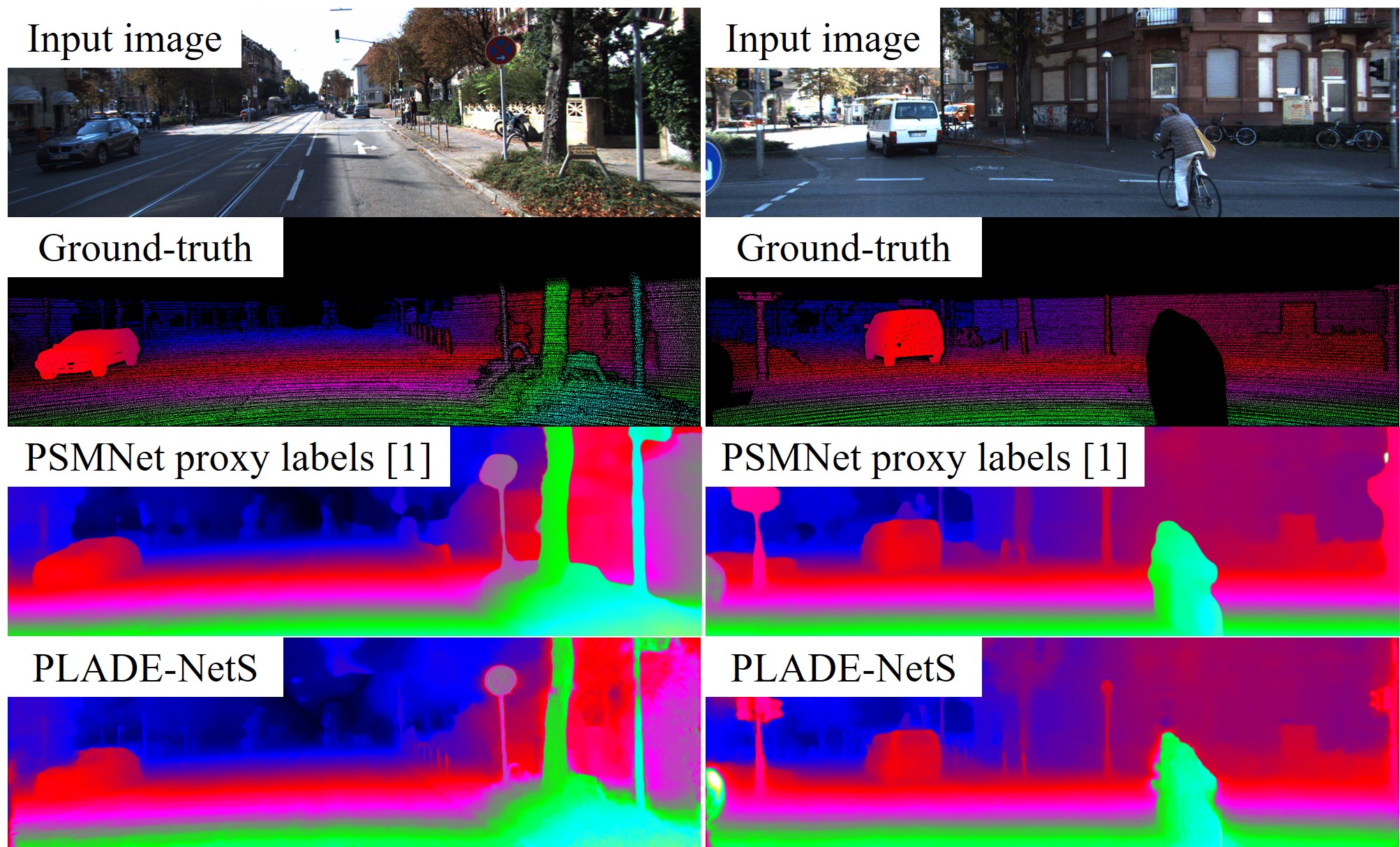}
  \vspace*{-7mm}
  \caption{Stereo results on the KITTI2015\cite{kitti2015} dataset.}
  \label{fig:kitti2015}
  \vspace*{-2mm}
\end{figure}

\begin{table}[t]
    \small
    \centering
    \setlength{\tabcolsep}{1.7pt}
    % \begin{tabular}{clccccccccccc}
% \hline
% Ref & Methods & PP & Sup & Data & \#Par & abs rel$\downarrow$  & sq rel$\downarrow$  & rmse$\downarrow$  & rmse$_{\text{log}}\downarrow$  & $a^1\uparrow$ & $a^2\uparrow$ & $a^3\uparrow$ \\ 
% \hline

% \cite{monodepth1} & Godard \etal & & S & K & 32 & 0.068 & 0.835 & 4.392 & 0.146 & 0.942 & 0.978 & 0.989 \\
% \cite{bridging}   & Lai \etal & & S & K & - & 0.062 & 0.747 & 4.113 & 0.146 & 0.948 & 0.979 & 0.990 \\
% \cite{unos}       & UnOS (stereo-only) & & S & K & - & 0.060 & 0.833 & 4.187 & 0.135 & 0.955 & 0.981 & 0.990 \\
% \cite{unos}       & UnOS & & SV & K & - & \underline{0.049} & 0.515 & 3.404 & 0.121 & 0.965 & 0.984 & 0.992 \\
% \cite{reversing}  & PSMNet with proxy & & S & K & - &  - &  - & 3.764 & 0.115 & \textbf{0.974} & 0.988 & 0.993 \\
% our               & PLADE-Nets & & S & K & 23 & 0.053 & \underline{0.323} & \underline{2.758} & \underline{0.100} & 0.965 & \underline{0.989} & \underline{0.995} \\
% our               & PLADE-Nets & \bluecheck & S & K & 23 & \textbf{0.050} & \textbf{0.300} & \textbf{2.723} & \textbf{0.096} & \underline{0.967} & \textbf{0.990} & \textbf{0.996} \\

% \hline
% \end{tabular}

\begin{tabular}{lccccccc}
\hline
Method & abs rel\cellcolor{c_lowbest}  & sq rel\cellcolor{c_lowbest}  & rmse\cellcolor{c_lowbest}  & rmse$_{\text{log}}$\cellcolor{c_lowbest}  & $\delta^1$\cellcolor{c_highbest} & $\delta^2$\cellcolor{c_highbest} & $\delta^3$ \cellcolor{c_highbest}\\ 
\hline

Godard \etal \cite{monodepth1} & 0.068 & 0.835 & 4.392 & 0.146 & 0.942 & 0.978 & 0.989 \\
Lai \etal \cite{bridging} & 0.062 & 0.747 & 4.113 & 0.146 & 0.948 & 0.979 & 0.990 \\
UnOS\cite{unos} & 0.060 & 0.833 & 4.187 & 0.135 & 0.955 & 0.981 & 0.990 \\
UnOS\cite{unos} (SV) & \underline{0.049} & 0.515 & 3.404 & 0.121 & 0.965 & 0.984 & 0.992 \\
Aleotti \etal \cite{reversing} & - &  - & 3.764 & 0.115 & \textbf{0.974} & 0.988 & 0.993 \\
PLADE-NetS & 0.053 & \underline{0.323} & \underline{2.758} & \underline{0.100} & 0.965 & \underline{0.989} & \underline{0.995} \\
PLADE-NetS & \textbf{0.050} & \textbf{0.300} & \textbf{2.723} & \textbf{0.096} & \underline{0.967} & \textbf{0.990} & \textbf{0.996} \\
\hline
\end{tabular}
    \vspace*{-2mm}
    \caption{Comparison of existing self-supervised SDE methods on the KITTI2015 \cite{kitti2015} training set. SV: Training from stereo videos. \textbf{Best} and \underline{second-best} metrics. Results capped to 80m.}
    \label{tab:kitti_2015}
    \vspace*{-3mm}
\end{table}

\begin{figure*}
  \centering 
  \includegraphics[width=0.97\textwidth]{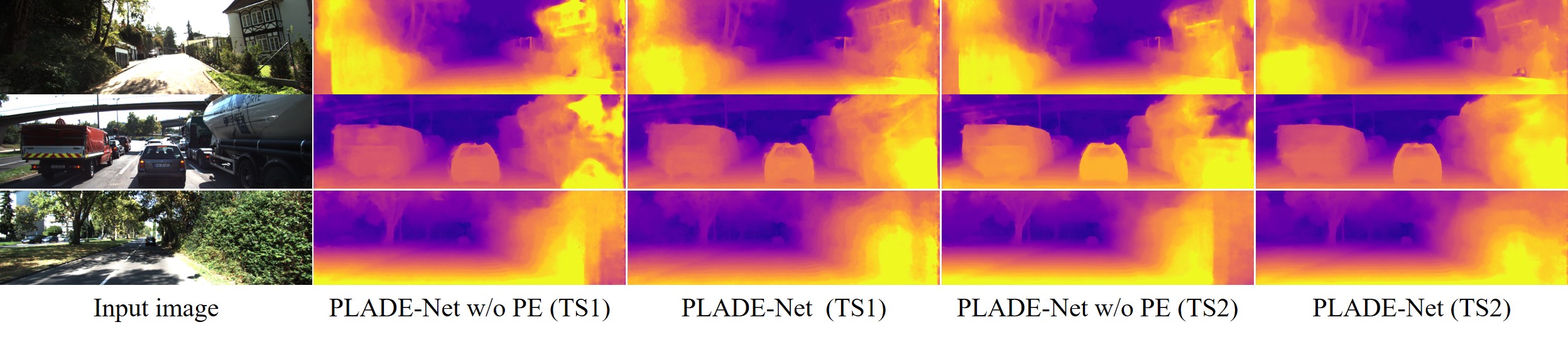}
  \vspace*{-4mm}
  \caption{Qualitative ablation studies on our proposed neural positional encoding (NPE) in our PLADE-Net.}
  \label{fig:ablation_npe}
  \vspace*{-4mm}
\end{figure*}

\subsection{Ablation Studies}
Table \ref{tab:ablation} shows our ablation studies on the improved KITTI Eigen test split \cite{kitti_official}. We first ablate the effects of our NPE in the PLADE-Net for the first stage of training. As it can be noted, the positional encoding (PE) that simply concatenates the pixel location $(x, y)$ values directly into the encoder stage can even yield slight performance improvements in most metrics in comparison with the cases without it (denoted as \enquote{w/o PE}). However, our PLADE-Net shows substantial performance improvements in all metrics by incorporating our \textit{neural} positional encoding (NPE).

The ablation studies on the effects of our NPE in the second stage of training are shown in the second section of Table \ref{tab:ablation}. Interestingly, our PLADE-Net w/o PE gets stuck in bad local minima, with marginal performance improvements in the second training stage. In contrast, our PLADE-Net with simple PE outperforms all previous SOTA methods in terms of $\delta^1$ accuracy. On the other hand, our PLADE-Net with NPE exhibits the best performance by considerable margins. Our PLADE-Net with NPE trained only with the KITTI Eigen train split \cite{eigen} shows the efficacy of robust learning regardless of the training data size. The effects of our NPE in both training stages (TS1 and TS2) are depicted in Figure \ref{fig:ablation_npe}. Our PLADE-Net without NPE struggles to estimate depths for objects close to the image borders, yielding depth artifacts regardless of the training stage.

The third section of Table \ref{tab:ablation} and Figure \ref{fig:loss_ablation} show respectively the quantitative and qualitative ablation studies for our proposed loss functions. As can be noted, our distilled matting Laplacian loss is the main contributor to achieving such high-performance. Our PLADE-Net trained with distilled matting Laplacian loss only ($a_{dc}=0$, $a_{dm}=0.25$) achieves an $\delta^1$ accuracy of 94.8\% while our PLADE-net with deep corr-$l_1$ loss only ($a_{dc}=0.01$, $a_{dm}=0$) obtains the lower accuracy of 94.5\%. In addition, training with only deep corr-$l_1$ loss induces depth artifacts seen as bright spots in Figure \ref{fig:loss_ablation}-(d). Our PLADE-Net without our proposed loss functions ($a_{dc}=0$, $a_{dm}=0$) shows the lowest performance with the most blur depth estimates as shown in Figure \ref{fig:loss_ablation}-(c). Interestingly, our PLADE-Net with ($a_{dc}$=$0$, $a_{dm}$=$0$) still shows considerably better metrics than the previous SOTA FAL-net \cite{falnet}, demonstrating the effectiveness of our NPE and design choices of multi-scale inputs and more learned features on the shallow feature extractors, suggesting that SVDE benefits from richer low-level features.

\begin{figure*}
  \centering 
  \includegraphics[width=1.0\textwidth]{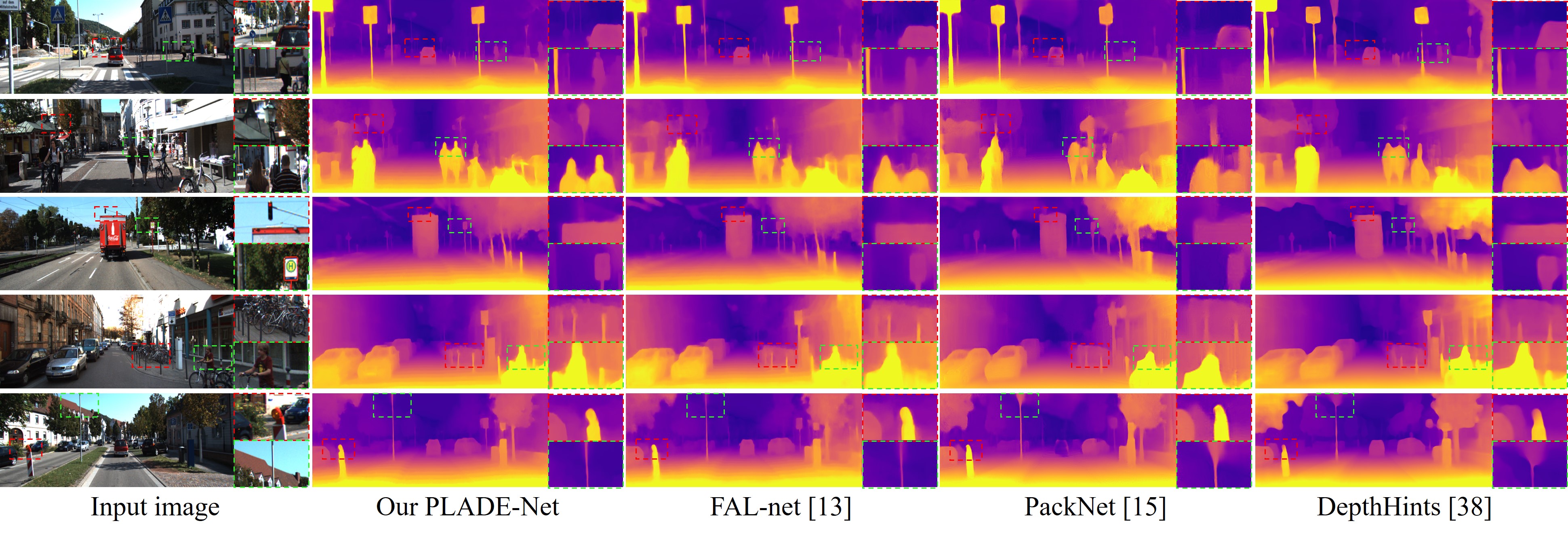}
  \vspace*{-9mm}
  \caption{Qualitative comparisons on the KITTI Eigen test split \cite{eigen}. Our PLADE-Net consistently estimates much more detailed depths.}
  \label{fig:res_eigen}
\end{figure*}

\subsection{Resutls}
\textbf{Results on KITTI.} Table \ref{tab:kitti_eigen} and Figure \ref{fig:res_eigen} present quantitative and qualitative comparisons among the previous methods and our PLADE-Net on the KITTI Eigen test split \cite{eigen}. Our PLADE-Net clearly outperforms all previous self-supervised methods in most metrics on the original Eigen test split \cite{eigen}, and in all the metrics on the improved test split \cite{kitti_official}. Our PLADE-Net shows sharper and pixel-level accurate depth estimates in complex and cluttered image regions, as shown in every zoom-box of Figure \ref{fig:res_eigen}. Quantitatively, our PLADE-Net without any post-processing (PP), even trained only on the KITTI (K) dataset, outperforms the previous methods that were trained on KITTI + CityScapes (K+CS). Following the PP step in \cite{falnet}, our method achieves even higher accuracies and lower error metrics. 

\begin{table*}[t]
    \small
    \centering
    \begin{tabular}{clccccccccccc}
\hline
Ref & Methods & PP & Sup & Data & \#Par & abs rel\cellcolor{c_lowbest} & sq rel\cellcolor{c_lowbest}  & rmse\cellcolor{c_lowbest}  & rmse$_{\text{log}}$\cellcolor{c_lowbest}  & $\delta^1$\cellcolor{c_highbest} & $\delta^2$\cellcolor{c_highbest} & $\delta^3$\cellcolor{c_highbest} \\ 
\hline

\rowcolor{c_data}
\multicolumn{13}{c}{Original Eigen Test Split \cite{eigen}}\\
% Supervised
\cite{defocus}          & Gur \etal & & DoF & K & - & 0.110 & 0.666 & 4.186 & \underline{0.168} & 0.880 & 0.966 & \textbf{0.988} \\ % Depth of field					
\cite{singleviewstereo} & Luo \etal & & D+S  & K & - & 0.094 & 0.626 & 4.252 & 0.177 & 0.891 & 0.965 & 0.984 \\	
\hline
				
% Self-sup video
% \cite{depthwild}    & Gordon \etal & & V & K & - & 0.128 & 0.959 & 5.230 & 0.212 & 0.845 & 0.947 & 0.976 \\
\cite{monodepth2}   & Monodepth2 & & V & K & 14 & 0.115 & 0.882 & 4.701 & 0.190 & 0.879 & 0.961 & 0.982 \\	
\cite{packing3d}    & PackNet & & V & K & 120 & 0.107 & 0.802 & 4.538 & 0.186 & 0.889 & 0.962 & 0.981 \\
\cite{depthwild}    & Gordon \etal & & V & K+CS & - & 0.124 & 0.930 & 5.120 & 0.206 & 0.851 & 0.950 & 0.978 \\
\cite{packing3d}    & PackNet & & V & CS$\rightarrow$K & 120 & 0.104 & 0.758 & 4.386 & 0.182 & 0.895 & 0.964 & 0.982 \\ 
\cite{semguide}     & Guizilini \etal & & V+Se & CS$\rightarrow$K & 140 & 0.100 & 0.761 & 4.270 & 0.175 & 0.902 & 0.965 & 0.982 \\
\hline

% Self-sup stereo
% \cite{monodepth1}     & Monodepth &             & S & K & 32 & 0.148 & 1.344 & 5.927 & 0.247 & 0.803 & 0.922 & 0.964 \\
\cite{superdepth}     & SuperDepth &            & S & K & - & 0.112 & 0.875 & 4.958 & 0.207 & 0.852 & 0.947 & 0.977 \\
\cite{infuse_classic} & Tosi \etal & \redcheck & S$_{\text{SGM}}$ & K & 42 & 0.111 & 0.867 & 4.714 & 0.199 & 0.864 & 0.954 & 0.979 \\
\cite{refinedistill}  & Refine\&Distill &         & S & K & - & 0.098 & 0.831 & 4.656 & 0.202 & 0.882 & 0.948 & 0.973 \\
\cite{depth_hints}    & DepthHints & \redcheck   & S$_{\text{SGM}}$ & K & 35 & 0.096 & 0.710 & 4.393 & 0.185 & 0.890 & 0.962 & 0.981 \\
% \cite{monodepth1}     & Monodepth & \redcheck    & S & CS$\rightarrow$K & 32 & 0.114 & 0.898 & 4.935 & 0.206 & 0.861 & 0.949 & 0.976 \\
\cite{net3}           & 3Net & \redcheck         & S & CS$\rightarrow$K & 48 & 0.111 & 0.849 & 4.822 & 0.202 & 0.865 & 0.952 & 0.978 \\
\cite{infuse_classic} & Tosi \etal & \redcheck & S$_{\text{SGM}}$ & CS$\rightarrow$K & 42 & 0.096 & 0.673 & 4.351 & 0.184 & 0.890 & 0.961 & 0.981 \\
% \cite{falnet} & FAL-net &            & S & K & 17 & 0.099 & 0.633 & 4.074 & 0.177 & 0.894 & 0.965 & 0.984} \\
\cite{falnet} & FAL-net & \redcheck & S & K & 17 & 0.094 & 0.597 & 4.005 & 0.173 & 0.900 & 0.967 & 0.985 \\
% \cite{falnet} & FAL-net &            & S & K+CS & 17 & 0.091} & 0.562} & 4.016 & 0.178 & 0.894 & 0.964 & 0.983 \\
\cite{falnet} & FAL-net & \redcheck & S & K+CS & 17 & \underline{0.088} & \textbf{0.547} & 4.004 & 0.175 & 0.898 & 0.966 & 0.984 \\
\hline

% Ours
our & PLADE-Net &            & S & K & 15 & 0.092 & 0.626 & 4.046 & 0.175 & 0.896 & 0.965 & 0.984 \\
our & PLADE-Net & \redcheck & S & K & 15 & 0.089 & 0.590 & 4.008 & 0.172 & 0.900 & 0.967 & 0.985 \\
our & PLADE-Net &            & S & K+CS & 15 & 0.090 & 0.577 & \underline{3.880} & 0.170 & \underline{0.903} & \underline{0.968} & \underline{0.985} \\
our & PLADE-Net & \redcheck & S & K+CS & 15 & \textbf{0.087} & \underline{0.550} & \textbf{3.837} & \textbf{0.167} & \textbf{0.908} & \textbf{0.970} & \underline{0.985} \\
% 0.091 & 0.553 & 3.979 & 0.177 & 0.894 & 0.965 & 0.984 (normal pp at same scale)
\hline

% Trained on CS eval on K
% \cite{depthwild}  & Gordon \etal & & V & CS & - & 0.172 & 1.370 & 6.210 & 0.250 & 0.754 & 0.921 & 0.967 \\
% \cite{falnet}     & FAL-net   & & S & CS & 17 & \textit{0.144} & \textit{0.871} & \textit{4.796} & \textit{0.215} & \textit{0.811} & \textit{0.947} & \textit{0.979} \\
% our             & PLADE-Net   & & S & CS & 15 &  \\
% \hline

\rowcolor{c_data}
\multicolumn{13}{c}{Improved Eigen Test Split \cite{kitti_official}}\\
% Supervised
% \cite{singleviewstereo} & Luo \etal & & DS  & K & - &  \\	
\cite{dorn}       & DORN & & D & K & 51 & 0.072 & 0.307 & 2.727 & 0.120 & 0.932 & 0.984 & 0.995 \\
\hline

% Self-sup video
\cite{monodepth2} & Monodepth2 & & V & K & 14 & 0.092 & 0.536 & 3.749 & 0.135 & 0.916 & 0.984 & 0.995 \\
\cite{packing3d}  & PackNet (LR) & & V & K & 120         & 0.078 & 0.420 & 3.485 & 0.121 & 0.931 & 0.986 & 0.996 \\
\cite{packing3d}  & PackNet & & V & CS$\rightarrow$K & 120 & 0.071 & 0.359 & 3.153 & 0.109 & 0.944 & 0.990 & 0.997 \\ 
\hline

% Self-sup stereo
% \cite{monodepth2}  & Monodepth2 &            & V+S & K & 14 & 0.087 & 0.479 & 3.595 & 0.131 & 0.916 & 0.984 & 0.996 \\	
\cite{monodepth2}  & Monodepth2 &            & S & K & 14 & 0.084 & 0.503 & 3.646 & 0.133 & 0.920 & 0.982 & 0.994 \\	
\cite{depth_hints} & DepthHints & \redcheck & S$_{\text{SGM}}$ & K & 35 & 0.074 & 0.364 & 3.202 & 0.114 & 0.936 & 0.989 & 0.997 \\
\cite{falnet}      & FAL-net & \redcheck & S & K & 17 & 0.071 & 0.281 & 2.912 & 0.108 & 0.943 & 0.991 & 0.998 \\
\cite{falnet}      & FAL-net & \redcheck & S & K+CS & 17 & 0.068 & 0.276 & 2.906 & 0.106 & 0.944 & 0.991 & 0.998 \\
\hline

% Ours
our & PLADE-Net &  & S & K & 15 & 0.066 & 0.274 & 2.881 & 0.105 & 0.944 & 0.992 & 0.998 \\
our & PLADE-Net & \redcheck & S & K & 15 & 0.066 & 0.272 & 2.918 & 0.104 & 0.945 & 0.992 & 0.998\\
our & PLADE-Net & & S & K+CS & 15 & \underline{0.066} & \underline{0.263} & \underline{2.726} & \underline{0.102} & \underline{0.949} & \textbf{0.992} & \textbf{0.998} \\
our & PLADE-Net & \redcheck & S & K+CS & 15 & \textbf{0.065} & \textbf{0.253} & \textbf{2.710} & \textbf{0.100} & \textbf{0.950} & \textbf{0.992} & \textbf{0.998}\\
\hline
our & PLADE-NetS & & S & K & 15 & 0.036 & 0.094 & 1.791 & 0.060 & 0.988 & 0.998 & 0.999\\
our & PLADE-NetS & \redcheck & S & K & 15 & 0.035 & 0.091 & 1.748 & 0.058 & 0.989 & 0.998 & 1.000\\
\hline
\end{tabular}
    \vspace*{-2mm}
    \caption{Evaluations on the KITTI Eigen test split \cite{eigen}. Models are trained on the KITTI Eigen\cite{eigen} train-split (K) and CityScapes\cite{cityscapes} (CS).  CS$\rightarrow$K indicates CS pre-training. K+CS indicates concurrent K and CS training. DoF and D denote depth-of-field and depth supervision. S, S$_{\text{SGM}}$, V, V+Se indicate stereo, stereo+SGM, video, and video + semantics self-supervision. V methods benefit from median-scaling. \textbf{Best} and \underline{second-best} metrics. Methods that use post-processing (PP) are checked \redcheck. Results capped to 80m.}
    \label{tab:kitti_eigen}
    \vspace*{-7mm}
\end{table*}

\textbf{Results on KITTI (stereo)}
To further evaluate the effectiveness of our PLADE-Net, we define an stereo input variant, the PLADE-NetS, which is evaluated and compared with the SOTA methods on the KITTI2015 dataset. Our stereo variant is a clone of the PLADE-net, with the difference that the PLADE-NetS \enquote{naively} incorporates the right-view image information in a second encoder, whose bottleneck features are concatenated to the left view bottleneck features. In our PLADE-NetS we do not incorporate any advanced stereo matching layers such as 1D-Correlation or 3D convolutions, and still, our network manages to outperform the most recent self-supervised SOTA methods \cite{reversing, unos} in most metrics by a considerable margin, as indicated in Table \ref{tab:kitti_2015}. Figure \ref{fig:kitti2015} shows that our network with stereo inputs keeps generating very sharp and pixel-level accurate depth estimates with clear object boundaries.

\textbf{Results on Make3D.} Table \ref{tab:make3d} compares our PLADE-Net against the SOTA self-supervised methods on Make3D \cite{make_3d}. Our approach generalizes the best among the self-supervised methods under comparison and is very close to the fully-supervised method of Laina \etal \cite{lainadeeper}. It is clear in Figure \ref{fig:make_3d} that our PLADE-Net generates sharper depth estimates on the previously unseen Make3D dataset \cite{make_3d} in comparison with the recent FAL-net \cite{falnet} SOTA. 

\section{Conclusions}
We showed that our PLADE-Net with neural positional encoding (NPE) could generalize better than the conventional CNN approaches. NPE allows our PLADE-Net to learn location-specific features, which aid in predicting consistent disparities in all image regions. Furthermore, our proposed distilled matting Laplacian loss provides strong self-supervision signals to learn sharp and pixel-level accurate depth estimation. Our PLADE-Net outperforms all previous self-, semi-, and fully-supervised methods on the challenging KITTI dataset with unprecedented accuracy levels and exhibits superior generalization capacities on the Make3D and CityScapes datasets. 

% Furthermore, our PLADE-NetS outpeforms the most recent self-supervised SDE methods without any advanced layer for stereo matching.

{\small
\bibliographystyle{ieee_fullname}
\bibliography{egbib}
}

\end{document}